\documentclass{article}

\usepackage[nonatbib,preprint]{neurips_2022}
\usepackage[numbers]{natbib}
\usepackage[utf8]{inputenc}
\usepackage[T1]{fontenc}
\usepackage{hyperref}
\usepackage{url}
\usepackage{booktabs}
\usepackage{amsfonts}
\usepackage{nicefrac}
\usepackage{microtype}
\usepackage{xcolor}

\usepackage{mathtools}
\DeclarePairedDelimiter{\norm}{\lVert}{\rVert}
\usepackage{amsmath}
\usepackage{amssymb}
\usepackage{amsthm}
\usepackage{algpseudocode}
\usepackage{algorithm}
\theoremstyle{definition}
\newtheorem{definition}{Definition}
\newcounter{claims}[section]
\newenvironment{claim}[1][]{\refstepcounter{claims}\par\noindent\underline{Claim~\theclaims:}\space#1}{}
\newenvironment{claimproof}[1]{\par\noindent\underline{Proof:}\space#1}{\hfill $\blacksquare$}
\let\originalparagraph\paragraph
\renewcommand{\paragraph}[2][.]{\originalparagraph{#2#1}}

\newcommand{\argmin}{\mathop{\mathrm{argmin}}\limits}
\newcommand{\argmax}{\mathop{\mathrm{argmax}}\limits}

\title{Meet You Halfway: Explaining Deep Learning Mysteries}

\author{ Oriel BenShmuel 
\\
Faculty of Math\&CS \\
Weizmann Institute of Science \\
Israel \\
\texttt{oriel.benshmuel@weizmann.ac.il}
}
\date{\vspace{-5ex}}

\begin{document}

\date{\vspace{-5ex}}

\maketitle

\begin{abstract}
Deep neural networks perform exceptionally well on various learning tasks with state-of-the-art results. While these models are highly expressive and achieve impressively accurate solutions with excellent generalization abilities, they are susceptible to minor perturbations. Samples that suffer such perturbations are known as ``adversarial examples''. Even though deep learning is an extensively researched field, many questions about the nature of deep learning models remain unanswered. In this paper, we introduce a new conceptual framework attached with a formal description that aims to shed light on the network's behavior and interpret the behind-the-scenes of the learning process. Our framework provides an explanation for inherent questions concerning deep learning. Particularly, we clarify: (1) Why do neural networks acquire generalization abilities? (2) Why do adversarial examples transfer between different models?. We provide a comprehensive set of experiments that support this new framework, as well as its underlying theory.

\end{abstract}

\section{Introduction}
Past work had successfully promoted deep neural networks to be one of the top-list models when it comes to learning tasks such as classification, prediction, and generation of new information \cite{voulodimos2018deep,redmon2016you,he2017mask,ren2015faster,goodfellow2014generative,vaswani2017attention,devlin2018bert,bahdanau2014neural}. Many problems in computer vision and natural language processing have reached state-of-the-art results using deep neural networks \cite{grill2020bootstrap,tolstikhin2021mlp,tan2020efficientdet,dosovitskiy2020image,liu2021swin,mildenhall2020nerf,brown2020language,lan2019albert,dai2019transformer}. Naturally, these models attracted a lot of attention in the artificial intelligence community. Along with their profound impact, deep neural networks raise many questions regarding their behavioral nature.

Adversarial examples are modified samples based on minor perturbations that are hardly imperceptible to the human visual system \cite{szegedy2013intriguing}. Yet, these perturbations cause even a well-trained network to misclassify the produced image. Furthermore, the nature of adversarial examples is not associated with any specific network. The same adversarial example would manipulate (with high probability) a different network trained with a different architecture and disjoint dataset \cite{szegedy2013intriguing}. The latter phenomenon is known as ``transferability'', and currently, its causes remain unclear \cite{liu2017delving,tramer2017space}. Naturally, the ``transferability'' occurrence might be used to generate black-box attacks, where one might generate an adversarial example without using any information regarding the target model \cite{liu2017delving}. 

A series of studies discussed this disadvantage with various efforts for its prevention and for the creation of robust models \cite{madry2017towards,tramer2020adaptive}. Several approaches aimed to explain the nature of adversarial perturbations. Two notable examples are the ``centroid'' approach and the ``robust and non-robust features'' theory \cite{ilyas2019adversarial,kim2021distilling}. The “centroid” approach, while not officially defined, generally refers to the adversarial direction as some weighted direction toward the closest samples from the target class (for example, it was mentioned by Ian Goodfellow in his lecture \cite{goodfellowtalk} at 1:11:54). The second approach \cite{ilyas2019adversarial} refers to ``robust features'' and ``non-robust features''. The robust and non-robust features theory, on the other hand, defines robust features as patterns that are predictive of the true label even when adversarially perturbed. Conversely,  non-robust features correspond to patterns that, while predictive, can be ``flipped” by an adversary within a predefined perturbation set to indicate a wrong class.

Another series of theoretical studies explore the ability of deep neural networks to generalize \cite{kawaguchi2017generalization,neyshabur2017exploring,chuang2021measuring}. Despite the non-convex nature of the optimization problem, simple optimization methods such as SGD have been found to perform well \cite{li2018learning,cao2019generalization}. The broad set of possible parameters attached to the training process constitutes the possible converged states of the model. This over-parametrized setting may offer many possible solutions for the objective, but not all of them demonstrate a well-generalized solution \cite{ying2019overview}. Optimization with respect to the training set is partially contradictory to achieving ample performance on the test set, resulting in a bias-variance tradeoff. In other words, the goal is to learn the training set distribution without overfitting. Multiple approaches have been suggested to alleviate this problem \cite{kukavcka2017regularization,zhang2021understanding}. 

Despite these conceptual ideas and practical suggestions, some of the fundamental questions concerning deep neural networks, such as ``Why do networks generalize?'' or ``Why do adversarial examples transfer?'' are still vague. In this paper, we explain the evolution of the decision boundary and its properties during the network's learning process. In particular, we expose an additional optimization goal of the training. This reveal provides us with a set of straightforward explanations of the intrinsic and counter-intuitive occurrences ``generalization'' and ``transferability''. We begin in sections \ref{mental} and \ref{adver_dir} by illustrating the properties of the decision boundary and explaining under what constraints it evolves (with emphasis on the new optimization property we expose). Next, in sections \ref{transfer} and \ref{gen}, we unfold the two phenomena described earlier through the lens of the elaborated optimization properties. Finally, in section \ref{exp}, we provide a broad set of experiments to support our claims. For the sake of simplicity, we consider only 2-class classifiers and use $\ell_2$ norms.

\section{Paradigm shift}
\label{mental}

Generally, a set of representatives sampled from the data distribution might be separated in multiple independent manners according to the provided labels. However, the separation itself is not the only consideration of a classification network. Deep neural networks are constrained to a set of optimization requirements when learning the data, which lead to the desired goal of correct classification. In this paper, we mention these constraints and describe one novel requirement of the optimization process. 

Apart from the optimization requirements, our new framework illustrates the learning process as a difference-based mechanism, where mutual properties of the classes, such as objective background (unrelated to a particular class distribution), will be neglected in the network consideration (with the exception of statistical errors). In fact, we might find many features related to the semantic meaning of the classes, and yet, they will be neglected as they are distributed similarly for both classes. Therefore, these features will not contribute any new insights regarding the differences between the classes. Following this perspective, any feature that the network learns represents some property expressed dichotomously among the two classes. 

The optimization process aims to detect dichotomous properties that are not associated only to a specific separator but to the class distributions (as we clarify formally in subsection \ref{formal}). In other words, these dichotomous properties are based on the gap between the two class distributions. In that case, reduction of the entirety of these dichotomous properties would generate indistinguishable distributions of the two classes (subject to the limited accuracy of the settings). However, reduction of only some of these properties would only reduce the distributional difference between the classes. Assume a dataset that appropriately represents the classes' distributions, with enough samples to indicate some of the mentioned differences. Reduction of these indicative differences would be expressed with a Euclidean distance reduction between the classes, even when using complex datasets and natural images (for more details, see section \ref{exp}).

To simplify this concept, we define the \emph{projection} of a sample in the input space onto the decision boundary of a classification network to be the sample's nearest point on the decision boundary. We refer to \emph{global difference} as a measurement tool that indicates the extent to which the projected training set (with respect to the trained network's decision boundary) reduced the difference between the classes' distributions in a particular set of directions in the input space. In other words, after reducing a ``global difference'' property (by using the projection of the training set), it will be reduced with respect to any possible (separating) boundary. This intuitive mental image is formally explained in subsection \ref{formal} under the assumptions in subsection \ref{assumptions}.

\subsection{Formal description}
\label{formal}
\par
Let $x\in \mathbb{R}^n$ be an input sample, and $l\in\{0,1\}$ is the corresponding label. For a broader context, denote $\mathcal{I}=\{x_i\}_{i=1}^s$ as a set of samples from the data distribution with the corresponding labels $\mathcal{L}=\{l_i\}_{i=1}^s$. To address the problem more elegantly, we define (or redefine) some expressions to the scope of this paper.

\paragraph{Definition 1 (Classifier)}
The mapping \(f:\mathbb{R}^n\longrightarrow\mathbb{R}\) is a \emph{classifier} for the set of samples and labels \((\mathcal{I},\mathcal{L})\), if \(f(x_i)f(x_j)<0\) for any \(x_i, x_j \in \mathcal{I}\) such that \(l_i\ne l_j\). We denote \(\mathcal{C}_{(\mathcal{I},\mathcal{L})}\) as the set of all the classifiers for \((\mathcal{I},\mathcal{L})\).

\paragraph{Definition 2 (Decision boundary)}
The \emph{decision boundary} \(\mathcal{B}_f\subseteq \mathbb{R}^n\) of any classifier \(f\in \mathcal{C}_{(\mathcal{I},\mathcal{L})}\) is defined as \(\mathcal{B}_f \coloneqq f^{-1}(0)\).

\paragraph{Definition 3 (Projection)}
The \emph{projection} \(\mathcal{P}_{\mathcal{B}_f}:\mathbb{R}^n\longrightarrow\mathbb{R}^n\) of sample \(x\in\mathbb{R}\) onto the decision boundary  \(\mathcal{B}_f\) for \(f\in \mathcal{C}_{(\mathcal{I},\mathcal{L})}\) is defined as \(\mathcal{P}_{\mathcal{B}_f}(x) \coloneqq \argmin_{b\in\mathcal{B}_f} \norm{b-x}\) and the \emph{projection vector} as \(\overline{\mathcal{P}_{\mathcal{B}_f}}(x) \coloneqq \mathcal{P}_{\mathcal{B}_f}(x)-x\).

\noindent
\underline{Note:} In this paper (unless stated differently), we discuss decision boundaries with a well-defined function $\mathcal{P}_{\mathcal{B}_f}(x)$, such that exists only a single possible projection for each input. In section \ref{transfer}, we discuss cases with several possible projection sets for the given dataset $\mathcal{I}$ and decision boundary $\mathcal{B}_f$.

\begin{paragraph}{Definition 3 (Global difference)}
Assume a classifier \(f\in \mathcal{C}_{(\mathcal{I},\mathcal{L})}\), where the number of the samples is \(|\mathcal{I}|=s\), and the projected set is \(\mathcal{I'}=\{x'_i\}_{i=1}^s=\{\mathcal{P}_{\mathcal{B}_f}(x_i)\}_{i=1}^s\). The \emph{global difference} value of \(f\) w.r.t \((\mathcal{I},\mathcal{L})\) is:
$$\phi(f, \mathcal{I}, \mathcal{L}) \coloneqq s-\max\limits_{g\in\mathcal{C}_{(\mathcal{I'},\mathcal{L})}}  \bigg\{\sum \limits_{i=1}^s \alpha_i \bigg|\forall_{1\leq i\leq s}: \overline{\mathcal{P}_{\mathcal{B}_g}}(x'_i)=\alpha_i\overline{\mathcal{P}_{\mathcal{B}_f}}(x_i)\wedge 0\leq\alpha_i\leq1\bigg\}$$
\end{paragraph}

\begin{figure}[htp!]
     \centering
     \includegraphics[width=\textwidth]{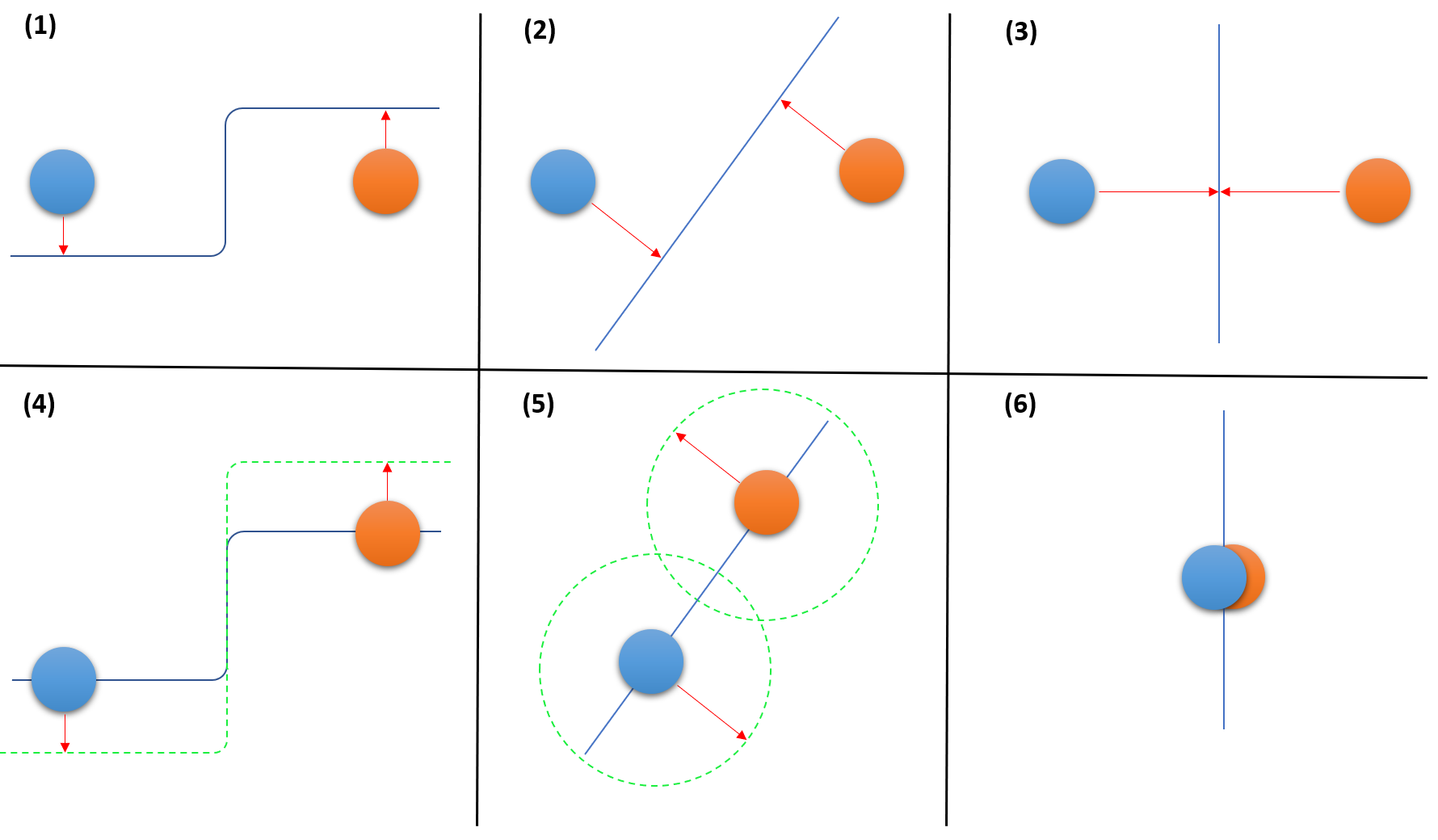}
     \caption{In the first row, there are different classifiers for the same problem. We can see the projections on the first row's boundaries in the second row. Each one of them has a different ``global difference'' value.}
     \label{fig:gaps}
\end{figure}

In Figure \ref{fig:gaps}, we can observe an example of different separators for the same problem in the first row (plots $1,2,3$) and the corresponding projections sets in the second row (plots $4,5,6$). In plot $4$, we can see the projection of plot $1$, and as the green separator indicates, there is a separator that could satisfy $\overline{\mathcal{P}_{\mathcal{B}_g}}(x')=\overline{\mathcal{P}_{\mathcal{B}_f}}(x)$ for all the samples. Plot $5$ is the projection of plot $2$, and as noted by the green circles, we could not separate the samples while generating similar projection vectors (with the same lengths). Instead, we could use $\overline{\mathcal{P}_{\mathcal{B}_g}}(x')=\alpha\overline{\mathcal{P}_{\mathcal{B}_f}}(x)$ for some $\alpha<1$ to create a separator with the same directions. In plot $3$, we generated a separator that puts the examples as far as possible from the decision boundary. In that case, we could not generate a separating boundary for the projected set (hence $\alpha=0$), as illustrated in plot $6$.

Our proposed approach for explaining the intrinsic properties of neural networks claims that the learning process aims to generate a classifier that satisfies the following constraint for any representative training set $(\mathcal{I},\mathcal{L})$ (limited to the expressive power of the architecture), a claim which we support by informal evidence and quantitative experiments in section \ref{exp}:
$$f_\phi(\mathcal{I},\mathcal{L})=\argmax_{f\in \mathcal{C}_{(\mathcal{I},\mathcal{L})}}\phi(f, \mathcal{I}, \mathcal{L})$$

Among all the possible solutions for $f_\phi(\mathcal{I},\mathcal{L})$, the network would prefer a classifier that places each training example as far as possible from its decision boundary to maximize the confidence level in its provided labels subject to the limited expressive power of the given DNN. Therefore, we narrow down the possible solutions to the following:
$$F_\phi(\mathcal{I},\mathcal{L})=\argmax\limits_{f\in f_\phi(\mathcal{I},\mathcal{L})} \prod\limits_{i=1}^s \norm{\overline{\mathcal{P}_{\mathcal{B}_f}}(x_i)}$$

\subsection{Assumptions, limitations, and clarifications}
\label{assumptions}

Here are some clarifications regarding the new approach:
\begin{enumerate}
    \item To avoid contradiction with the primary preference of the network (to classify the samples correctly), and due to the limited complexity of the network, a single projection of the training set onto the decision boundary will not reduce (in most natural cases) all of the global differences at once.

    \item Our approach describes an optimization problem that aims to reach a \emph{global property}, and it is not necessarily reflected when observing each sample separately.

    \item Assume the process of reducing global differences by projecting the training set iteratively (project the samples, find a new separator, project the new samples, etc.). Our definition of ``global difference'' is sensitive to the iterations' order. Applying the same set of reductions in a different order could not occur according to our maximization requests.

    \item We assume that the network is randomly initialized at the beginning of the training process, and the initialized state is not pathologically close to any specific solution (that might not meet the new constraints).

    \item The claimed behavior ignores statistical deviations and numerical errors that might occur in some datasets and perform unexpected solutions.

    \item Due to the network's degree limitations, the regularization parameters (preventing highly expressive behavior near the samples), and the interest of the decision boundary to stay far from the samples (as described in section \ref{formal}), we assume that the decision boundary does not change its direction frequently and drastically in small regions and specifically near the samples.

\end{enumerate}

\section{Euclidean distance and adversarial directions}
\label{adver_dir}
As shown in section \ref{exp}, projecting the dataset might be expressed with a Euclidean distance reduction between the classes' representatives in the input space. This puts into question whether our definition of ``global difference'' is based on the Euclidean distance between the samples. The concrete notion of ``global difference'' is not directly related to a Euclidean distance but rather the (possibly) nonlinear properties that differentiate the two classes. However, reducing the gap between the class distributions for a well-representative dataset sampled from the data distribution (particularly for each class) will motivate a Euclidean distance reduction. Generally, this metric reduction will be expressed globally (in average terms) rather than individually for each training set sample.

The ``centroid'' approach refers to a reduction (namely, the adversarial direction) toward the closest samples from the target class. In contrast, our framework claims for an entirely different occurrence. In our case, samples' adversarial directions are not influenced by specific samples, but rather all of them are influenced globally. Specifically, they are not influenced due to local regions' representatives but rather by a global property that might be expressed differently in different local regions. Moreover, we might observe the similarity between the approaches in very simple and low-dimensional cases. In that case, the adversarial directions of close points might be related, not as a result of their individual interaction, but rather due to the similar behavior of the decision boundary in that specific region. While the ``centroid'' approach might perform well on elementary problems, this is not the case for complex and natural datasets (where the samples are usually very far from each other in the first place). These complex cases are explained clearly with our framework, which we support with experiments (for more details, see section \ref{exp}).

\section{Transferability}
\label{transfer}

In 2013, Szegedy et al. presented the new idea of adversarial examples and some of their essential properties \cite{szegedy2013intriguing}. In the same paper, two additional properties of adversarial examples were introduced:
\begin{enumerate}
\item\emph{Cross model generalization:} Adversarial examples will be misclassified, with high probability, by networks trained with different architecture and hyper-parameters.
\item\emph{Cross training-set generalization:}
Adversarial examples will be misclassified, with high probability, by networks trained with disjoint datasets (but similar semantic meaning).
\end{enumerate}

These properties are also known as the ``transferability'' occurrence.

In this section, we aim to explain the ``transferability'' occurrence. In particular, we claim that our proposed optimization approach for natural datasets generates a single global minimum, and any other minimum is relatively negligible and performs an inferior solution with an improbable convergence process compared to the primary solution. In contrast, datasets with a few meaningful solutions (based on our approach) will have a symmetrical structure, and it is most probable that transferability will not occur among networks that converge to different solutions. We assume that natural datasets are not symmetrically structured due to the possibly unrelated distribution of the background and other natural properties.

\par
\begin{definition}[Symmetrical dataset]
The dataset \(\mathcal{I}\) with the corresponding labels \(\mathcal{L}\) is a \emph{symmetrical dataset} if, for any solution \(f\in F_\phi(\mathcal{I},\mathcal{L})\), we could generate a few possible different projection sets \(\{\mathcal{P}_{\mathcal{B}_f}(x)\}_{x\in \mathcal{I}}\) for the same classifier \(f\).
\end{definition}

\begin{figure}[htp!]
     \centering
     \includegraphics[width=\textwidth]{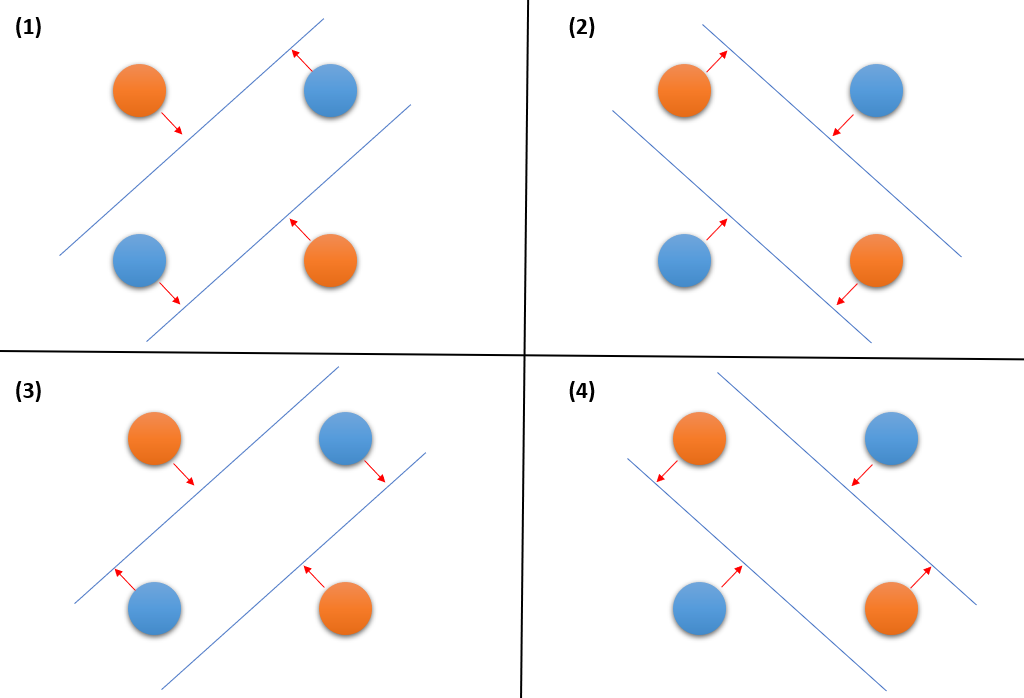}
     \caption{In each column, there are different sets of projections for the same decision boundary.}
     \label{fig:symm}
\end{figure}

In Figure \ref{fig:symm}, we can observe two different decision boundaries for the same problem. Each decision boundary enables several options for a valid projection.

Assume a pair of possible independent solutions $f,g\in F_\phi(\mathcal{I},\mathcal{L})$ such that:
$$\forall x\in\mathcal{I}: \overline{\mathcal{P}_{\mathcal{B}_f}}(x)\perp\overline{\mathcal{P}_{\mathcal{B}_g}}(x)$$

In Appendix \ref{appendix:transfer_unqueness}, we prove that the described solutions $f,g$ could not co-exist for a non-symmetrical dataset.

Overall, the optimization process of networks that learn natural (and non-symmetrical) datasets has a unique global minimum.
Therefore, different networks with different settings that converge to the described global minimum (subject to the architecture's constraints) will probably provide a similar solution that is expressed with a similar decision boundary. According to \cite{tramer2017space}, decision boundary similarity enables ``transferability'' as claimed. Additionally, networks that perform well on the dataset with different expressiveness abilities might reach different optimization scores even though it is most probable they will converge to the (same) global minimum. Therefore their decision boundary might look different in some local regions, but the global behavior will stay the same (including transferable adversarial examples).

\section{Generalization}
\label{gen}
A deep neural network trained with a representative dataset will be able (for relevant settings) to generalize its correct classification with high probability, also when presented with samples that did not participate in the training set. This occurrence is known as \emph{Generalization}. In this section, we propose a simple explanation for the described effect through the lens of our framework.

In section \ref{formal}, we defined the term ``global difference''. In the described notation, we encouraged the optimization process to find a classifier that maximizes the following expression:
$$\max\limits_{g\in\mathcal{C}_{(\mathcal{I},\mathcal{L})}}  \bigg\{\sum \limits_{i=1}^s \alpha_i \bigg|\forall_{1\leq i\leq s}: \overline{\mathcal{P}_{\mathcal{B}_g}}(x'_i)=\alpha_i\overline{\mathcal{P}_{\mathcal{B}_f}}(x_i)\wedge 0\leq\alpha_i\leq1\bigg\}$$
Following that expression, we limit the value of $\alpha_i$ for each example $i$. Additionally, we assume that the decision boundary is as far as possible from the samples under the described conditions. This approach motivates the optimization process to involve a significant fraction of the training set in order to maximize the expression. For a representative training set, we get that the solutions for $F_\phi(\mathcal{I},\mathcal{L})$ are directly related to the data distribution. Moreover, as described in section \ref{mental} and supported by section \ref{exp}, the solutions for $F_\phi(\mathcal{I},\mathcal{L})$ are based on a difference between the classes which is objectively unique through the lens of any classifier (and cannot be reduced twice). Therefore it is not exclusively related to the general distribution of the data but also to the gap between the individual distributions of the classes. 

According to our definition of $F_\phi(\mathcal{I},\mathcal{L})$, the generated classifier is the one that capable of performing the largest reduction of the gap between the class distributions using the most significant distributional differences among them. As explained, it would tend to find global properties expressed with a large fraction of the dataset to satisfy this requirement. Overfitting occurs when these (significant) differences are already considered, and to further increase the reduction, the decision boundary starts to consider individual properties of specific distributional outliers. Therefore, regularization parameters and appropriate settings would only suppress the latest phase (of overfitting), where an overfitting solution is not inherently different from a generalizing solution but instead contains its global properties (and more).

Overall the detected solution separates the classes based on a distributional distinction. It implies that it would classify correctly, with high probability, other samples from the same distribution (for example, the test set).

\section{Experimental results}
\label{exp}
This section will experimentally analyze the reduction of ``global differences'' based on Euclidean distance between the two classes.

\subsection{Framework}
We aim to iteratively project the training set onto the decision boundary as described in Algorithm \ref{alg:iterproj}. In each iteration, we train a randomly initialized network using the produced training set. Then we project the training set of the current iteration onto the generated decision boundary of the trained network to replace the old training set with a new one (the projected one). In each iteration, we train the network until the accuracy reaches $90\%$ and all the samples are correctly classified. Additionally, for all iterations, we use an identical architecture and hyper-parameters to train the network, where the number of epochs might change (to reach the required accuracy).

\begin{algorithm}
\caption{Iterative projection}\label{alg:iterproj}
\begin{algorithmic}
    \While{iterations}
        \State model$\gets$Net()
        \State model.train($\mathcal{I}$, $\mathcal{L}$)
        \State $\mathcal{I}$ $\gets$ Project(model, $\mathcal{I}$)
    \EndWhile
\end{algorithmic}
\end{algorithm}

For each iteration, we measure the Euclidean distance between each sample of one class to its closest sample from the opposite class. Then, we compute the average of the resulted distances to indicate the distance between the classes for the current iteration. More formally, for a training set $(\{x_i\}_{i=1}^s,\{l_i\}_{i=1}^s)$ where $\forall_{1\leq i \leq s}:x_i\in\mathbb{R}^n,l_i\in\{0,1\}$, we compute for each iteration the following expression:
$$\frac{1}{s}\sum\limits_{i=1}^s{\min\limits_{1\leq j \leq s}\Big\{\norm{x_i-x_j} \Big| l_i\ne l_j\Big\}}$$

The analysis includes graphs with the described average distance for each iteration. Due to computational limitations, we randomly sampled a representative subset from the training set in each experiment (with equal classes ratio). We perform several experiments on a few different architectures for three datasets. The architectures for MNIST \cite{lecun2010mnist} and CIFAR10 \cite{CIFAR10} appear in Appendix \ref{appendix:architectures}, and for ImageNet, \cite{IMAGENET} we use resnet18 \cite{he2016deep} (using two output classes). We trained the networks using the negative log-likelihood loss function. The code was executed using \emph{CUDA 10.1} and \emph{Tesla K80 GPU}.

We show that the Euclidean distance between the classes is consistently reduced, which illustrates a trend rather than a random occurrence. We aim to illustrate the reduced gap between the class distributions. We use a Euclidean distance for this purpose while assuming that the gap between the class distributions is well-represented in the training set. In the experiments, we can see that the Euclidean distance reduced in earlier iterations is greater than the reductions that occur in later iterations. It implies the maximization property described in section \ref{formal} with ``global difference'' values and maximal distance of the examples from the decision boundary.

Following that, the ``error bar'' calculated using different seeds is unnecessary to our purpose in terms of the global trend as we initialized the network randomly in each iteration. Each segment in the presented graphs demonstrates similar behavior, even though we used hundreds of different initializations.

\subsection{Results}
\begin{figure}[htp!]
    \begin{minipage}[c]{0.5\linewidth}
    \centering
     \includegraphics[width=\textwidth]{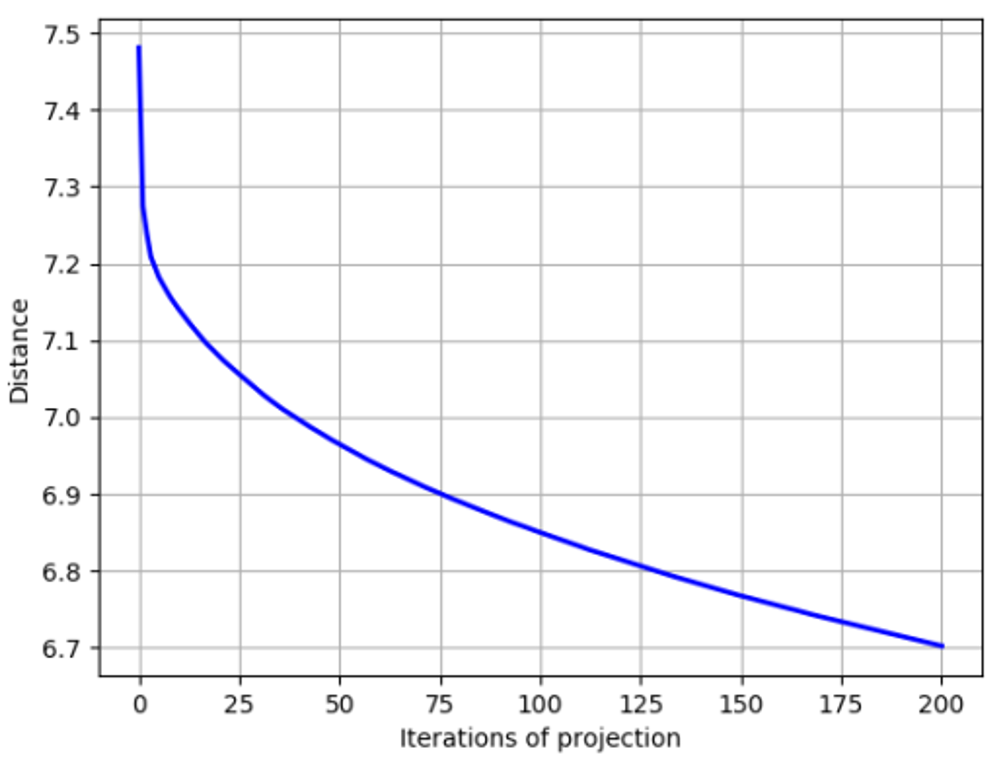}
     \caption{Global differences reduction process (MNIST).}
     \label{fig:mnist}
    \end{minipage}\hfill
     \begin{minipage}[c]{0.48\linewidth}
     \centering
      \includegraphics[width=\textwidth]{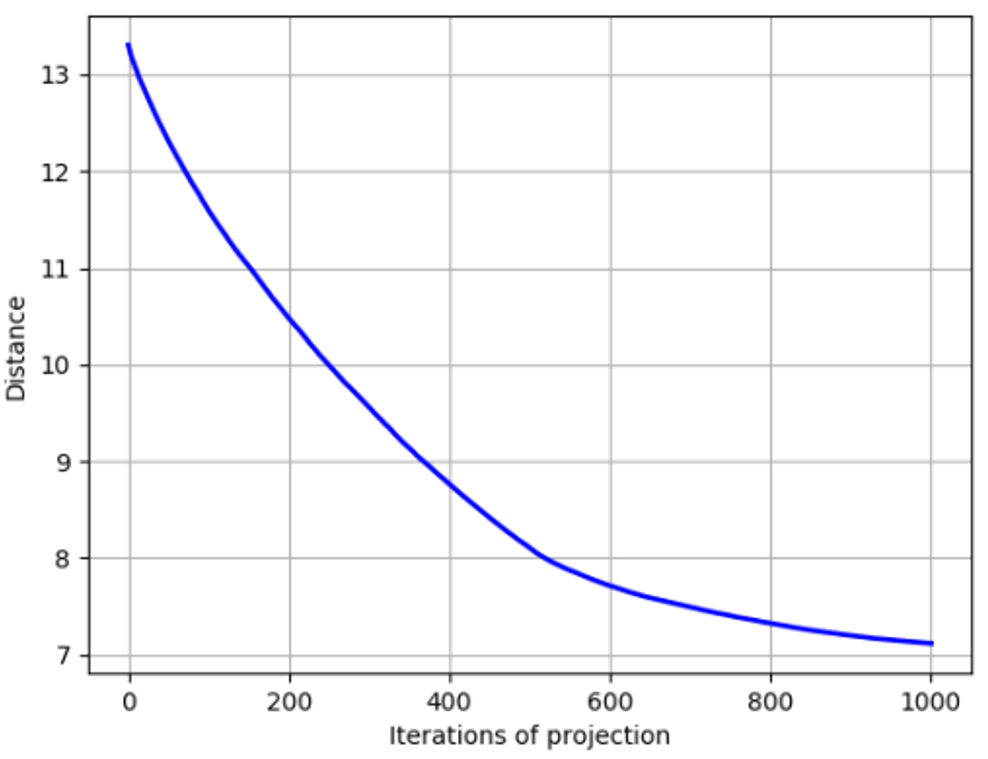}
     \caption{Global differences reduction process (CIFAR10).}
     \label{fig:cifar}
    \end{minipage}
\end{figure}

In Figure \ref{fig:mnist}, we demonstrate the analysis for $400$ samples from the MNIST \cite{lecun2010mnist} dataset, using the categories ``3'' and ``5'', Adam optimizer, and a learning rate of $1e-4$.

Figure \ref{fig:cifar} uses $200$ samples from the categories \emph{Airplane} and \emph{Automobile} of the CIFAR10 \cite{CIFAR10} dataset trained with Adam optimizer and a learning rate of $1e-4$.

\begin{figure}[htp!]
     \centering
     \includegraphics[width=\textwidth]{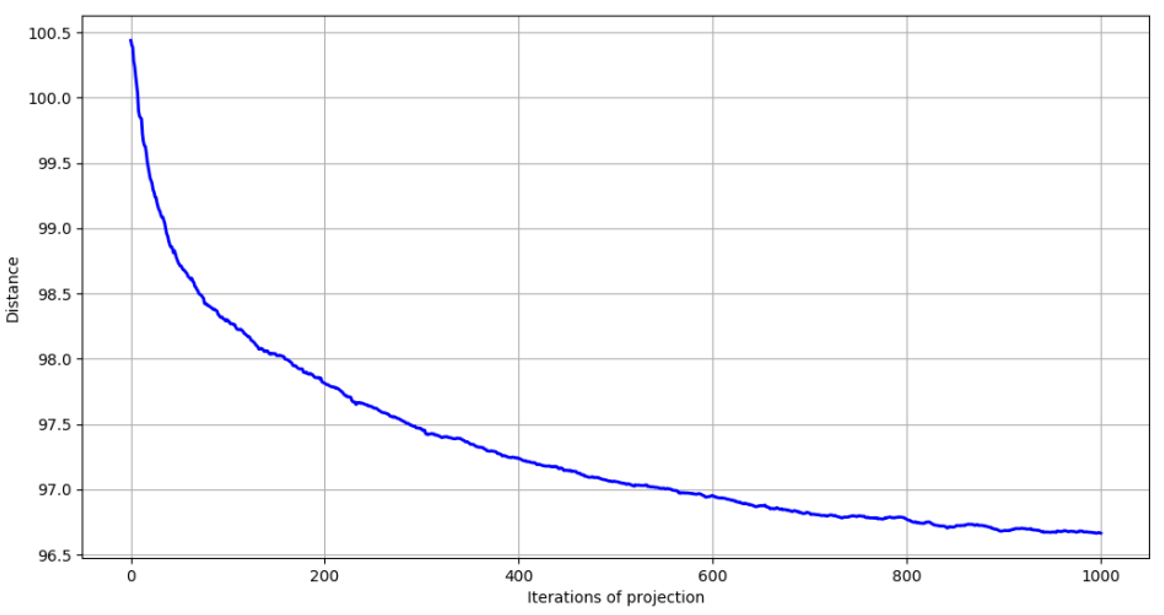}
     \caption{Global differences reduction process (ImageNet).}
     \label{fig:imgnet}
\end{figure}

Finally, Figure \ref{fig:imgnet} shows the analysis by randomly sampling $100$ samples from the categories \emph{Goldfish} and \emph{White shark} of the ImageNet \cite{IMAGENET} dataset. We used SGD with momentum and a learning rate of $1e-3$.

Additional experiments can be found in Appendix \ref{appendix:more_exps}.
\subsection{Limitations}
\label{exp_limits}

The experiments aim to show the properties we presented in section \ref{mental}. We address these properties as distributional characteristics. On the other hand, we base our experiments on the consistent reduction between the classes using Euclidean distance measurements. This detected gap aims to demonstrate the distributional difference between the class distributions.

Moreover, the experimental part states that earlier iterations show larger Euclidean distance reductions than the later ones. The claim is based on a global trend rather than individual observations, where for some specific iterations, we might get a smaller reduction than the one in the next iteration. We explain it by statistical deviations of the dataset, accuracy of only $90\%$, which might be expressed in a non-optimal decision boundary, and for later iterations, due to numerical errors and the limited expressiveness of the networks.

\section{Conclusion}
This work illustrates and formally presents an additional goal of the deep neural network's optimization process. This reveal assists us with the proposed explanations for some of the mysterious phenomena of deep learning. In particular, we explain the occurrences ``transferability'' and ``generalization'' through the lens of the presented framework. Finally, we provide support for this hypothesis by conducting novel experiments. The experiments show that the projection of the samples on the decision boundary, reduces the difference between the class distributions.

\bibliography{main}

\appendix
\section{Neural network architectures}
\label{appendix:architectures}

\begin{table}[htp!]
  
  \centering
  \caption{Neural network architectures used in this work.}\label{tab:arch}
  \begin{tabular}{ c c c c}
    \toprule
    MNIST & CIFAR10\\ 
    \midrule
    FC(784,500) + ReLU &  Conv(128,3,3) + BN + ReLU\\  
    FC(500,256) + ReLU &  Conv(128,3,3) + BN + ReLU\\  
    FC(256,128) + ReLU &  Conv(256,3,3) + BN + ReLU\\  
    FC(128,32) + ReLU &  MaxPool(2,2)\\ 
    FC(32,2) &  FC(1024,2)\\
    \bottomrule
  \end{tabular}
\end{table}

The architectures used in this work appear in Table \ref{tab:arch}. Conv: convolutional layer, FC: fully-connected layer, BN: batch normalization.

\section{Additional experiments}
\label{appendix:more_exps}
\begin{figure}[htp!]
     \centering
     \includegraphics[width=\textwidth]{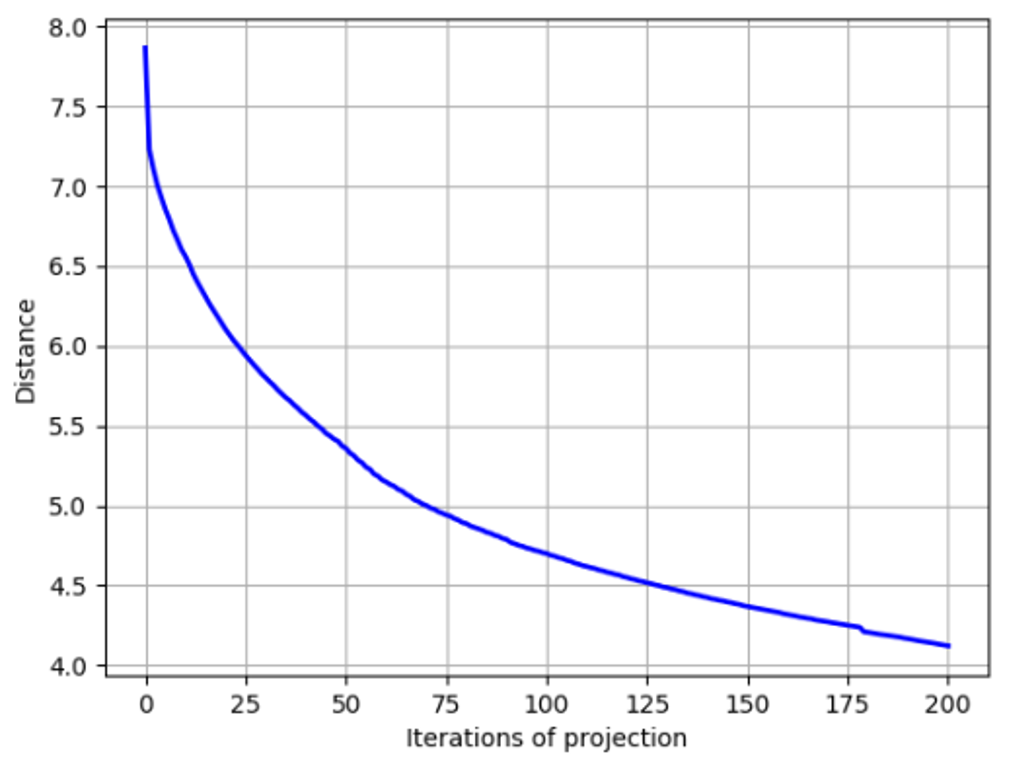}
     \caption{Global differences reduction process (``0'' versus ``1'' of MNIST).}
     \label{fig:mnist_add_exp}
\end{figure}

\begin{figure}[htp!]
     \centering
     \includegraphics[width=\textwidth]{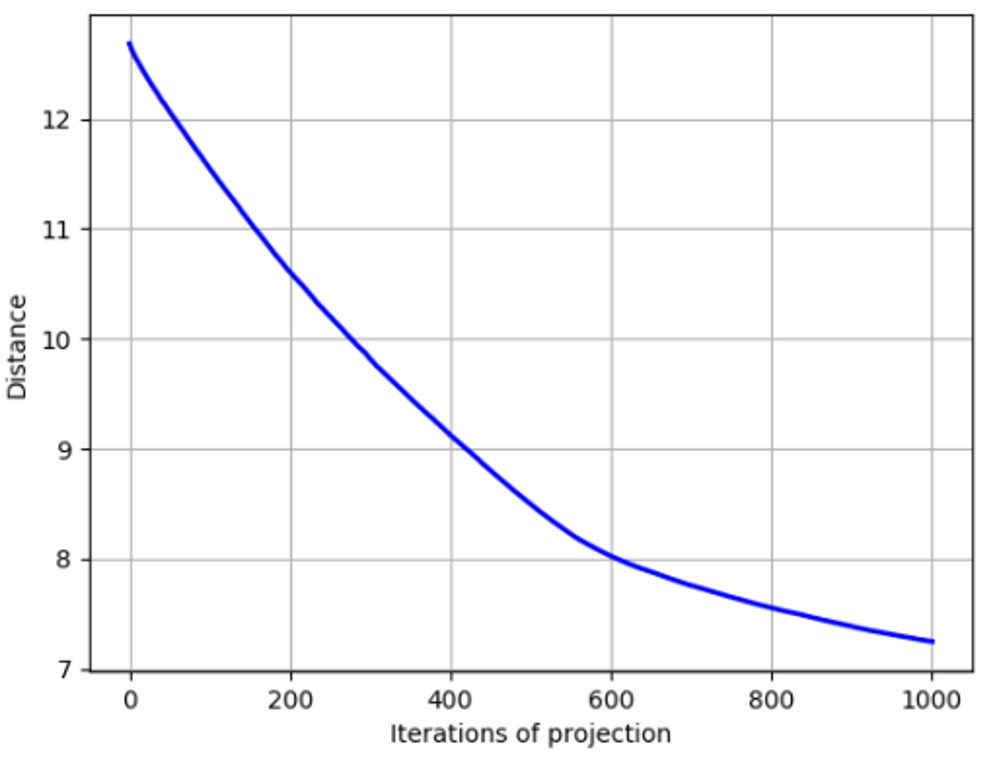}
     \caption{Global differences reduction process (\emph{Cat} versus \emph{Dog} of CIFAR10).}
     \label{fig:cifar_add_exp}
\end{figure}

\begin{figure}[htp!]
     \centering
     \includegraphics[width=\textwidth]{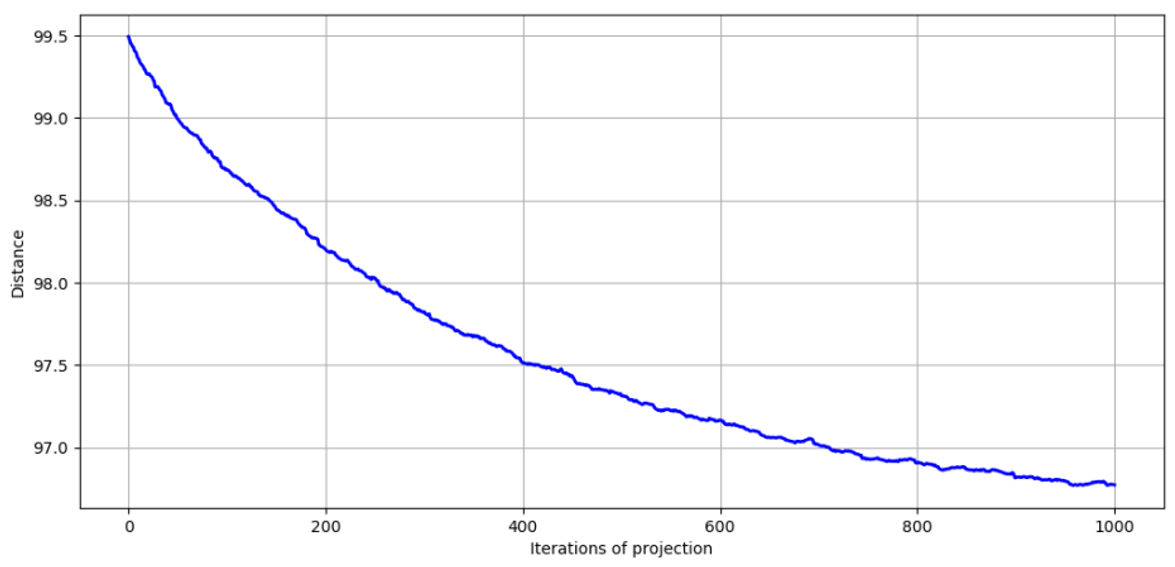}
     \caption{Global differences reduction process (\emph{Hen} versus \emph{Goose} of ImageNet).}
     \label{fig:imgnet_add_exp}
\end{figure}

We provide additional experiments with the same analysis for different categories. In Figures \ref{fig:mnist_add_exp},\ref{fig:cifar_add_exp},\ref{fig:imgnet_add_exp}, we analyze the classification of ``0'' versus ``1'' (MNIST), \emph{Cat} versus \emph{Dog} (CIFAR10), and \emph{Hen} versus \emph{Goose} (ImageNet), respectively.

\section{Solution uniqueness for non-symmetrical datasets}
\label{appendix:transfer_unqueness}

\begin{claim}
For a pair of possible independent solutions $f,g\in F_\phi(\mathcal{I},\mathcal{L})$ such that:
$$\forall x\in\mathcal{I}: \overline{\mathcal{P}_{\mathcal{B}_f}}(x)\perp\overline{\mathcal{P}_{\mathcal{B}_g}}(x)$$

the dataset $\mathcal{I}$ is symmetrical.
\end{claim}

\begin{claimproof}
According to prior definitions $f,g\in \mathcal{C}_{(\mathcal{I},\mathcal{L})}$. Therefore, for any $x_i,x_j\in\mathcal{I}$ with the corresponding labels $l_i,l_j\in\mathcal{L}$ such that $l_i\ne l_j$, the following must be applied:
$$\norm{\overline{\mathcal{P}_{\mathcal{B}_f}}(x_i)}+\norm{\overline{\mathcal{P}_{\mathcal{B}_f}}(x_j)}\leq\norm{x_i-x_j}$$
$$\norm{\overline{\mathcal{P}_{\mathcal{B}_g}}(x_i)}+\norm{\overline{\mathcal{P}_{\mathcal{B}_g}}(x_j)}\leq\norm{x_i-x_j}$$

Assume that the dataset is not symmetrical. Equality for $f$ could appear only when: $$\overline{\mathcal{P}_{\mathcal{B}_f}}(x_i)+\overline{\mathcal{P}_{\mathcal{B}_f}}(x_j)=x_i-x_j$$
and the same for $g$. Following the given assumption that $\overline{\mathcal{P}_{\mathcal{B}_f}}(x_i)\perp\overline{\mathcal{P}_{\mathcal{B}_g}}(x_i)$ and $\overline{\mathcal{P}_{\mathcal{B}_f}}(x_j)\perp\overline{\mathcal{P}_{\mathcal{B}_g}}(x_j)$, at least one of the above expressions is a strict inequality. Therefore we get:
$$\norm{\overline{\mathcal{P}_{\mathcal{B}_f}}(x_i)}+\norm{\overline{\mathcal{P}_{\mathcal{B}_f}}(x_j)}+\norm{\overline{\mathcal{P}_{\mathcal{B}_g}}(x_i)}+\norm{\overline{\mathcal{P}_{\mathcal{B}_g}}(x_j)}<\norm{x_i-x_j}+\norm{x_i-x_j}$$

$$\norm{\overline{\mathcal{P}_{\mathcal{B}_f}}(x_i)}+\norm{\overline{\mathcal{P}_{\mathcal{B}_f}}(x_j)}+\norm{\overline{\mathcal{P}_{\mathcal{B}_g}}(x_i)}+\norm{\overline{\mathcal{P}_{\mathcal{B}_g}}(x_j)}<2\norm{x_i-x_j}$$

$$\frac{1}{2}(\norm{\overline{\mathcal{P}_{\mathcal{B}_f}}(x_i)}+\norm{\overline{\mathcal{P}_{\mathcal{B}_g}}(x_i)})+\frac{1}{2}(\norm{\overline{\mathcal{P}_{\mathcal{B}_f}}(x_j)}+\norm{\overline{\mathcal{P}_{\mathcal{B}_g}}(x_j)})<\norm{x_i-x_j}$$

According to the triangle inequality:
$$\norm{\frac{1}{2}\big(\overline{\mathcal{P}_{\mathcal{B}_f}}(x_i)+\overline{\mathcal{P}_{\mathcal{B}_g}}(x_i)\big)}+\norm{\frac{1}{2}\big(\overline{\mathcal{P}_{\mathcal{B}_f}}(x_j)+\overline{\mathcal{P}_{\mathcal{B}_g}}(x_j)\big)}<\norm{x_i-x_j}$$

Following that, and the fact that $f,g\in \mathcal{C}_{(\mathcal{I},\mathcal{L})}$, we could generate a new classifier $h\in \mathcal{C}_{(\mathcal{I},\mathcal{L})}$ such that for any $x\in\mathcal{I}$ we define $\mathcal{P}_{\mathcal{B}_h}(x)=x+\frac{1}{2}\big(\overline{\mathcal{P}_{\mathcal{B}_f}}(x)+\overline{\mathcal{P}_{\mathcal{B}_g}}(x)\big)$.

In addition, Claim \ref{claim:mul} provides a proof for the following:
$$\prod\limits_{i=1}^s \norm{\overline{\mathcal{P}_{\mathcal{B}_h}}(x_i)}>\prod\limits_{i=1}^s \norm{\overline{\mathcal{P}_{\mathcal{B}_f}}(x_i)}=\prod\limits_{i=1}^s \norm{\overline{\mathcal{P}_{\mathcal{B}_g}}(x_i)}$$

We also assume that the ``global difference'' value of $f$ and $g$ is minimized. The classifier $h$ has the following projection $\mathcal{P}_{\mathcal{B}_h}(x)=x+\frac{1}{2}\big(\overline{\mathcal{P}_{\mathcal{B}_f}}(x)+\overline{\mathcal{P}_{\mathcal{B}_g}}(x)\big)$, with a direction that follows the directions of $\mathcal{P}_{\mathcal{B}_f}$ and $\mathcal{P}_{\mathcal{B}_g}$ for the specific point. Therefore, increasing the norm of the projection vectors in the same direction (using a larger scalar factor than $\frac{1}{2}$) will have to reach a minimized value of ``global difference'' as requested.

Overall, unless the dataset $\mathcal{I}$ is symmetrical, we could generate $h\in F_\phi(\mathcal{I},\mathcal{L})$ with $\prod\limits_{i=1}^s \norm{\overline{\mathcal{P}_{\mathcal{B}_h}}(x_i)}>\prod\limits_{i=1}^s \norm{\overline{\mathcal{P}_{\mathcal{B}_f}}(x_i)}=\prod\limits_{i=1}^s \norm{\overline{\mathcal{P}_{\mathcal{B}_g}}(x_i)}$ in contradiction.
\end{claimproof}

\begin{claim}
\label{claim:mul}
$\prod\limits_{i=1}^s \norm{\overline{\mathcal{P}_{\mathcal{B}_h}}(x_i)}>\prod\limits_{i=1}^s \norm{\overline{\mathcal{P}_{\mathcal{B}_f}}(x_i)}=\prod\limits_{i=1}^s \norm{\overline{\mathcal{P}_{\mathcal{B}_g}}(x_i)}$
\end{claim}

\begin{claimproof}
Based on the maximization property stated in section \ref{formal}:
$$\prod\limits_{i=1}^s \norm{\overline{\mathcal{P}_{\mathcal{B}_f}}(x_i)}=\prod\limits_{i=1}^s \norm{\overline{\mathcal{P}_{\mathcal{B}_g}}(x_i)}$$

According to our definition:
$\prod\limits_{i=1}^s \norm{\overline{\mathcal{P}_{\mathcal{B}_h}}(x_i)}=\prod\limits_{i=1}^s \norm{\frac{1}{2}\big(\overline{\mathcal{P}_{\mathcal{B}_f}}(x_i)+\overline{\mathcal{P}_{\mathcal{B}_g}}(x_i)\big)}$

After multiplying the elements, each operand in the summation has the following structure:
$$\frac{1}{2^s}\prod\limits_{i=1}^s\norm{p_i}$$
where $p_i\in\{\overline{\mathcal{P}_{\mathcal{B}_f}}(x_i),\overline{\mathcal{P}_{\mathcal{B}_g}}(x_i)\}$. For each operand like this, we also have the complement operand with  $p^c_i\in\{\overline{\mathcal{P}_{\mathcal{B}_f}}(x_i),\overline{\mathcal{P}_{\mathcal{B}_g}}(x_i)\}$ such that $p_i\ne p^c_i$ for any $i$.

Assume that for a specific operand, $T$ is the collection of indices where $p_i=\overline{\mathcal{P}_{\mathcal{B}_g}}(x_i)$. In that case, we could express the summation of the complement operands in the following manner:
$$\frac{1}{2^s}\prod\limits_{i=1}^s\norm{p_i}+\frac{1}{2^s}\prod\limits_{i=1}^s\norm{p^c_i}=$$

$$\frac{1}{2^s}(\prod\limits_{i=1}^s\norm{\overline{\mathcal{P}_{\mathcal{B}_f}}(x_i)})\cdot\frac{\prod\limits_{j\in T} \norm{\overline{\mathcal{P}_{\mathcal{B}_g}}(x_j)}}{\prod\limits_{j\in T}\norm{\overline{\mathcal{P}_{\mathcal{B}_f}}(x_j)}})+
\frac{1}{2^s}(\prod\limits_{i=1}^s\norm{\overline{\mathcal{P}_{\mathcal{B}_g}}(x_i)})\cdot\frac{\prod\limits_{j\in T} \norm{\overline{\mathcal{P}_{\mathcal{B}_f}}(x_j)}}{\prod\limits_{j\in T}\norm{\overline{\mathcal{P}_{\mathcal{B}_g}}(x_j)}})=$$

$$\frac{1}{2^s}(\prod\limits_{i=1}^s\norm{\overline{\mathcal{P}_{\mathcal{B}_f}}(x_i)})\cdot\frac{\prod\limits_{j\in T} \norm{\overline{\mathcal{P}_{\mathcal{B}_g}}(x_j)}}{\prod\limits_{j\in T}\norm{\overline{\mathcal{P}_{\mathcal{B}_f}}(x_j)}})+
\frac{1}{2^s}(\prod\limits_{i=1}^s\norm{\overline{\mathcal{P}_{\mathcal{B}_f}}(x_i)})\cdot\frac{\prod\limits_{j\in T} \norm{\overline{\mathcal{P}_{\mathcal{B}_f}}(x_j)}}{\prod\limits_{j\in T}\norm{\overline{\mathcal{P}_{\mathcal{B}_g}}(x_j)}})=$$

$$\frac{1}{2^s}\prod\limits_{i=1}^s\norm{\overline{\mathcal{P}_{\mathcal{B}_f}}(x_i)}\bigg(\frac{\prod\limits_{j\in T} \norm{\overline{\mathcal{P}_{\mathcal{B}_g}}(x_j)}}{\prod\limits_{j\in T}\norm{\overline{\mathcal{P}_{\mathcal{B}_f}}(x_j)}}+\frac{\prod\limits_{j\in T} \norm{\overline{\mathcal{P}_{\mathcal{B}_f}}(x_j)}}{\prod\limits_{j\in T}\norm{\overline{\mathcal{P}_{\mathcal{B}_g}}(x_j)}}\bigg)$$

For $a,b\in\mathbb{R}$ where $a,b>0$:
$$\frac{a}{b}+\frac{b}{a}\geq2$$
and we get equality only when $a=b$. Therefore, if $\prod\limits_{j\in T} \norm{\overline{\mathcal{P}_{\mathcal{B}_f}}(x_j)}\ne\prod\limits_{j\in T}\norm{\overline{\mathcal{P}_{\mathcal{B}_g}}(x_j)}$, we necessarily will get that:
$$\frac{1}{2^s}\prod\limits_{i=1}^s\norm{p_i}+\frac{1}{2^s}\prod\limits_{i=1}^s\norm{p^c_i}>\frac{2}{2^s}\prod\limits_{i=1}^s\norm{\overline{\mathcal{P}_{\mathcal{B}_f}}(x_i)}$$
Applying that for all the operands will generate the following inequality:
$$\prod\limits_{i=1}^s \norm{\overline{\mathcal{P}_{\mathcal{B}_h}}(x_i)}>\prod\limits_{i=1}^s \norm{\overline{\mathcal{P}_{\mathcal{B}_f}}(x_i)}$$

The same process could be applied to $g$.

In the case where $\prod\limits_{j\in T} \norm{\overline{\mathcal{P}_{\mathcal{B}_f}}(x_j)}=\prod\limits_{j\in T}\norm{\overline{\mathcal{P}_{\mathcal{B}_g}}(x_j)}$ and in particular for the cases where $|T|=1$, we get the following:
$$\forall_{1\leq i \leq s}:\norm{\overline{\mathcal{P}_{\mathcal{B}_f}}(x_i)}=\norm{\overline{\mathcal{P}_{\mathcal{B}_g}}(x_i)}$$

Combined with the earlier statement:
$$\frac{1}{2}(\norm{\overline{\mathcal{P}_{\mathcal{B}_f}}(x_i)}+\norm{\overline{\mathcal{P}_{\mathcal{B}_g}}(x_i)})+\frac{1}{2}(\norm{\overline{\mathcal{P}_{\mathcal{B}_f}}(x_j)}+\norm{\overline{\mathcal{P}_{\mathcal{B}_g}}(x_j)})<\norm{x_i-x_j}$$

we get that for any $i$ and $j$:
$$\norm{\overline{\mathcal{P}_{\mathcal{B}_f}}(x_i)}+\norm{\overline{\mathcal{P}_{\mathcal{B}_f}}(x_j)}<\norm{x_i-x_j}$$
In that case, we could generate $\alpha\norm{\overline{\mathcal{P}_{\mathcal{B}_f}}(x)}$ where $\alpha>1$ for at least one example while preserving the other properties and projection vectors. This is a contradiction to the maximization property of $f$. The same could have been claimed for $g$. Therefore, we got that $\prod\limits_{i=1}^s \norm{\overline{\mathcal{P}_{\mathcal{B}_h}}(x_i)}>\prod\limits_{i=1}^s \norm{\overline{\mathcal{P}_{\mathcal{B}_f}}(x_i)}=\prod\limits_{i=1}^s \norm{\overline{\mathcal{P}_{\mathcal{B}_g}}(x_i)}$.
\end{claimproof}

\end{document}